\documentclass[10pt,twocolumn,letterpaper]{article}

\usepackage{cvpr}
\usepackage{times}
\usepackage{epsfig}
\usepackage{graphicx}
\usepackage{amsmath}
\usepackage{amssymb}

\usepackage[linesnumbered,boxed]{algorithm2e}
\usepackage{color}
\usepackage{enumerate}

\usepackage{subfigure}

\usepackage{amsthm}

\theoremstyle{plain}

\theoremstyle{definition}

\usepackage[breaklinks=true,bookmarks=false]{hyperref}

\cvprfinalcopy 

\def\cvprPaperID{****} 

\begin{document}

\title{Light Field Segmentation From Super-pixel Graph Representation \thanks{The work was supported in part by NSFC under Grant 61531014 and Grant 61401359.}}

\author{Xianqiang Lv, Hao Zhu, Qing Wang\\
School of Computer Science\\
Northwestern Polytechnical University, Xi'an 710072, P.R. China\\
{\tt\small qwang@nwpu.edu.cn}
}

\maketitle
\thispagestyle{empty}

\begin{abstract}
   Efficient and accurate segmentation of light field is an important task in computer vision and graphics. The large volume of input data and the redundancy of light field make it an open challenge. In the paper, we propose a novel graph representation for interactive light field segmentation based on light field super-pixel (LFSP). The LFSP not only maintains light field redundancy, but also greatly reduces the graph size. These advantages make LFSP useful to improve segmentation efficiency. Based on LFSP graph structure, we present an efficient light field segmentation algorithm using graph-cuts. Experimental results on both synthetic and real datasets demonstrate that our method is superior to previous light field segmentation algorithms with respect to accuracy and efficiency.
\end{abstract}

\section{Introduction}

Light field camera is a powerful tool for capturing the 4D light field in a single shot. Compared with previous equipments, one of the most advantages of light field camera is passive depth estimation, which provides larger freedom for image segmentation in movie authoring industry. Due to the huge volume of 4D light field and the redundancy between different views, previous approaches \cite{Wanner2013Globally,mihara20164D} are either time-consuming or not proposed for full 4D data. On the other hand, image segmentation is a fundamental task in computer vision domain and a key step from low-level image processing to high-level image understanding. Accurate segmentation could be useful for face detection \cite{albiol2001unsupervised}, visual recognition \cite{shotton2006textonboost}, medical image \cite{grady2004multi} and so on. Since the regions of interest are different for different users or tasks, interactive segmentation is necessary to deal with this situation. In this paper, we develop a fast and accurate segmentation technique on full 4D light field data using the latest proposed light field super-pixel (LFSP) algorithm \cite{zhu20174d}.

\begin{figure}[t]
  \centering
  \includegraphics[width=1\linewidth]{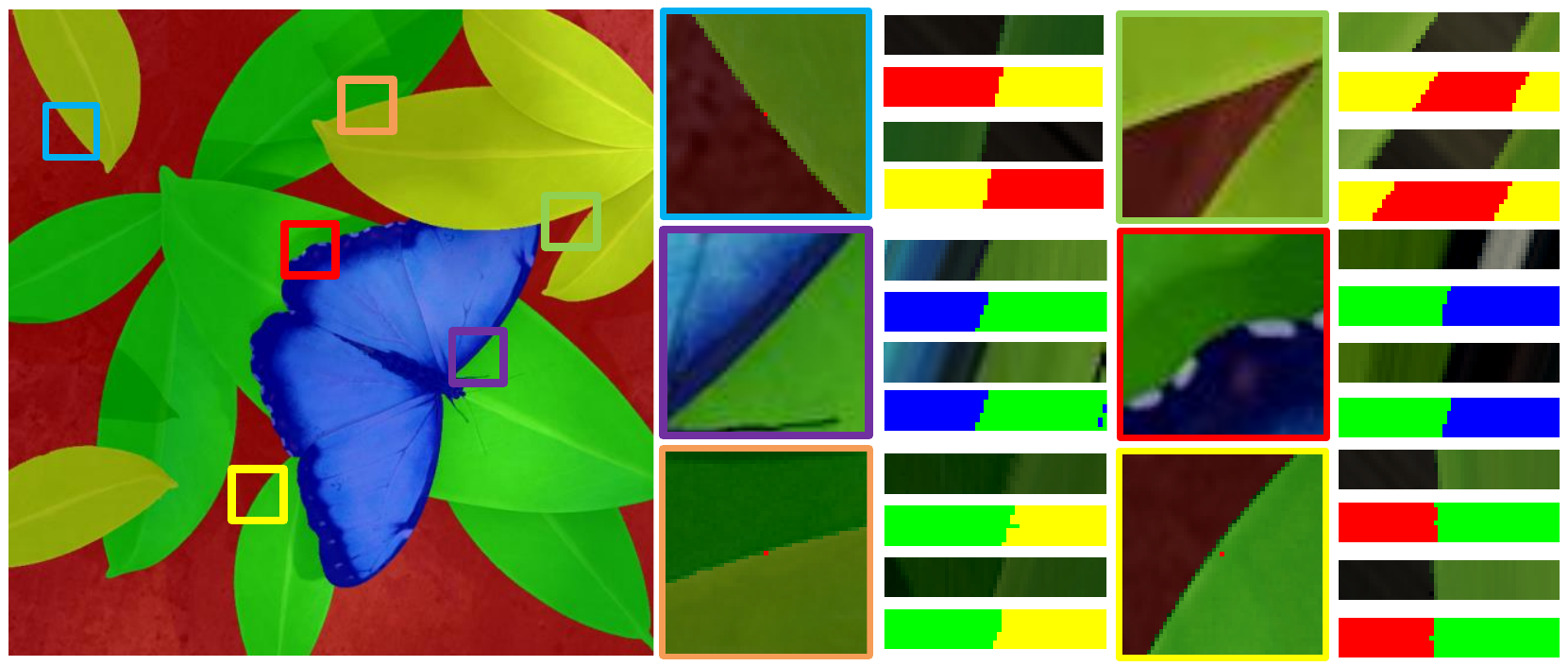}
  %
  \mbox{\small\hspace{12mm}(a)\hspace{18mm}(b)\hspace{8mm}(c)\hspace{8mm}(d)\hspace{8mm}(e)}\hfil

  \caption{\label{fig:im1}
  Light field segmentation results. (a) The overlay of raw image and segmentation result in the central view. (b,d) Close-ups of overlaying image. (c,e) From top to bottom: horizontal EPI of raw image, horizontal EPI of segmentation result, vertical EPI of raw image and vertical EPI of segmentation result respectively}
\end{figure}

Light field image records spatial and angular information of the scene by a 4D function named \(L(u,v,x,y)\), which has benefited many problems in computer vision, such as refocus \cite{ng2005fourier}, depth and scene flow estimation \cite{wanner2012globally,wang2015occlusion,zhu2017occlusion,srinivasan2015oriented}, material recognition \cite{wang20164d,zhang2016friction,Xue2017Differential} or super resolution \cite{bishop2012light,wanner2014variational}. By processing the ray recorded by light field, we could analyze the scene and solve some problems that could be intractable using traditional 2D image, such as segmentation ambiguity \cite{zhu20174d}. Some approaches \cite{Jarabo2011Efficient,Wanner2013Globally,mihara20164D} have been proposed to segment or edit light field, but the performance of these algorithms are not good enough because of two major difficulties mentioned by Jarabo et al. \cite{jarabo2014people}. First, the volume of 4D data leads to poor efficiency in terms of running time and memory consumption. Second, the segmentation result of different views should be coherent to preserve the redundancy between different views.

Light rays emitted from the region of a real object with similar characteristics are recorded and imaged by light field, which constitutes a LFSP. The LFSP can decrease the amount of data processing in spatial and angular domain. In position domain, the LFSP in each view contains many pixels from similar regions. In angle domain, the patch of each view is a part of a LFSP by fixing the angular dimension. Furthermore, since the 2D slice of the LFSP in each view records the light rays emitted from the same region, the LFSP remains coherent between different views and helps preserve the redundancy.

In this paper, we propose a novel light field graph structure based on LFSP. The users can interactively add labels on the central view image to specify the object of interest. The LFSP segmentation results and the user labels make up vertex of the graph. Then an energy function fusing position, appearance and disparity \cite{Wanner2013Globally} information is established and optimized using the graph cuts \cite{Boykov2002Fast}.

The experiments on both synthetic and real light field data captured by Lytro \cite{lytro_web} and ILLUM \cite{illum_web} demonstrate the effectiveness of the proposed algorithm. Quantitative comparisons show that the performance of our method in terms of accuracy and robustness  is competitive with the state-of-the-art methods. Moreover, thanks to its lower computational complexity, our method could be useful for some real-time applications.Our main contributions are summarized as follows,

(1) We propose a novel graph-based light field segmentation algorithm using LFSP. To the best of our knowledge, it is the first time such a solution using LFSP on full 4D light field data is available.

(2) Our method is capable of achieving competitive segmentation performance at a much lower running time complexity.

\section{Related Work}

As a fundamental problem in computer vision, segmentation has been extensively studied in recent decades. Various solutions are proposed to tackle the problem, such as region based segmentation \cite{salembier1999region}, threshold based segmentation \cite{lie1993efficient}, graph based segmentation \cite{Felzenszwalb2004Efficient}, learning based segmentation \cite{Dong2005Color} etc. Among these methods, the graph-cut algorithm \cite{boykov2001fast,Boykov2002Fast,kolmogorov2004energy,Felzenszwalb2004Efficient}, which has been proved to be an effective multi-label segmentation method, is closely related to our work. For graph-cut approaches, a graph structure consisting of vertices and edges is built to find a cut on the graph that minimizes the amount of energy by min-cut/max-flow algorithm \cite{boykov2004experimental}. Although these segmentation methods have good performance in specific conditions, a thorough segmentation solution solving the ambiguity in defocus and occlusion boundary areas is still an open challenge.

Light field \cite{Levoy1996Light} records angular and spatial information of scene through 4D data. Disparity \cite{Wanner2013Globally} that is closely related to depth-map is one of the most important features of light field. Several algorithms \cite{wanner2012globally,wang2015occlusion,zhu2017occlusion} have been proposed to generate an accurate disparity estimation. In this work, we use the algorithm developed by Zhu et al. \cite{zhu2017occlusion} to estimate the disparity. Furthermore, the structure of the scene could be analyzed through ray tracing, and this is an advantage for segmentation. However, the volume of 4D data and the redundancy of light field make light field segmentation extremely difficult \cite{jarabo2014people}.    As a result, segmentation approaches built on traditional 2D image are not suitable for 4D light field.

There are a few works about light field editing. In \cite{Jarabo2011Efficient}, the authors try to propagate the input edits in the full light field. They propose a novel multi-dimensional down-sampling and up-sampling techniques which are used to propagate the input edits. Although light field could be edited effectively to some extent, the results are greatly influenced by clustering. It is not good enough to deal with complex scenes. Jarabo et al. \cite{jarabo2014people} systematically analyze the ways of editing light field and construct corresponding interfaces, tools and workflows. This work gives detailed answers about how to edit light fields from a user perspective. Most of these methods rely on user inputs indicating the regions of interest to improve segmentation performance. In this paper, we need interactive user inputs to accurately segment complex scenes.

In recent years, several approaches are proposed to segment light field. Wanner et al. \cite{Wanner2013Globally} propose an algorithm for 4D light field segmentation for the first time, which introduces effective disparity cue. The authors use a set of input scribbles on the central view to train a random forest classifier based on disparity and appearance cues. Then, the global consistency of segmentation is optimized by using all views simultaneously. However, this method can only get the segmentation result of the central view. Moreover, it is time consuming. Xu et al. \cite{Xu2015TransCut} use the consistency and distortion properties of light field to segment transparent objects from background. However, the method is restricted to a certain type of background and the degree of reflection. Mihara et al. \cite{mihara20164D} define two neighboring ray types (spatial and angular) in light field data and design a learning-based likelihood function. Then they build a graph in 4D space and use graph-cut algorithm to get an optimized segmentation result. Since the complexity of their method is very high, only $5\!\times\!5$ view-points from $9\! \times\!9$ light field are used to reduce the data size in their experiment. For faster segmentation of light field, Hog et al. \cite{hog2016light} use a ray-based graph structure that exploits the redundancy in ray space to reduce the graph size, decreasing the running time of MRF-based optimization tasks. However, the proposed free-rays and ray bundles are greatly influenced by depth measurement. Moreover, the algorithm needs depth maps for all views, which are difficult to obtain.

Super-pixels are small regions consisting of a series of pixels similar to each other in position, appearance, brightness, etc. Most of these regions retain efficient information for further image segmentation. Generally, super-pixel do not destroy the boundary areas of an object. Many excellent algorithms \cite{Achanta2012SLIC,veksler2010superpixels} have been proposed for traditional 2D image. Zhu et al. \cite{zhu20174d} introduce light field super-pixel and LFSP self-similarity for the first time. Different from traditional super-pixel, LFSP is a 4D structure and keeps the redundancy in angular dimensions, which is the key characteristic of LFSP. Hog et al. \cite{hog17super} propose super-rays for efficient light field processing which is similar to LFSP to some extent.

\begin{figure*}[tbp]
  \centering
  \includegraphics[width=1\linewidth]{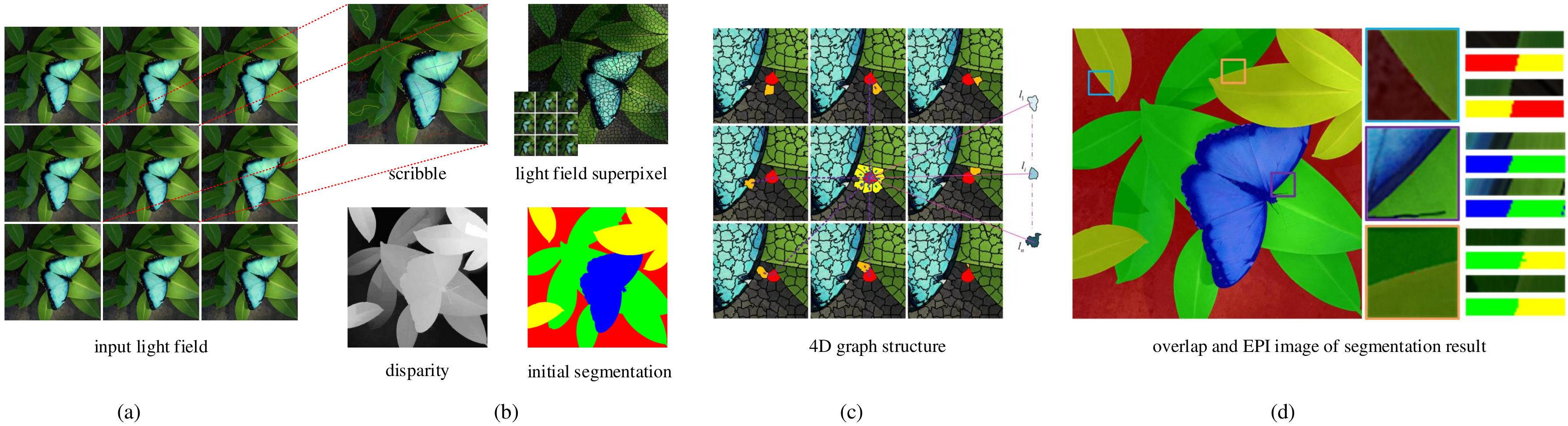}

  \caption{\label{fig:im2}%
           The pipeline of LFSP based light field segmentation. (a) $3 \times 3$ representation of input light field. (b) Data preprocessing and intermediate results, including user scribbles, disparity estimation, LFSP segmentation by \cite{zhu20174d} and initial segmentation results respectively. (c) 4D graph structure based on LFSP. (d) Optimized light field segmentation results.}
\end{figure*}

In contrast to previous approaches, we build a novel graph structure based on LFSP and treat LFSP as a unit of processing. Our algorithm can segment 4D light field as a whole and improve the accuracy and efficiency of light field segmentation. In the next sections, we will describe the graph model and the energy function respectively to cope with the above mentioned problems.

\section{Segmentation Based on LFSP}

In this section, we first introduce the representation of light field and the definition of LFSP. Then a graph structure based on LFSP is built. Finally, we formulate light field segmentation problem as an energy minimization problem.

The pipeline of the proposed algorithm is shown in Figure \ref{fig:im2}. The input contains a 4D light field and user scribbles in the central viewpoint (arbitrary viewpoint). Then, the disparity map for central view image is estimated, which benefits the LFSP segmentation. Next, the LFSP is calculated by Zhu et al. \cite{zhu20174d}. At the same time, the feature of each LFSP is initialized. Based on scribbles, disparity and LFSP, the initial segmentation result of light filed is obtained. Finally, a 4D graph structure is built and the optimized light field segmentation is achieved using graph-cuts. We use overlap and EPI images to show the optimized segmentation results.

\subsection{Representation of LFSP}

For describing physical world more objectively and realistically, Adelson and Bergen \cite{adelson1991plenoptic} propose a $7D$ plenoptic function. Light field is the result of reducing the dimension of plenoptic function, using $(u,v,s,t)$ to record a light ray. A light ray could be described by two-parallel-plane (TPP) model \cite{Levoy1996Light}. Supposing $uv$ plane is the view plane and $xy$ plane is the image plane. A ray $R_p$ emitted from a scene can be uniquely determined by $p=(u,v,x,y)$. Another way of representing light field is the multi-view representation, which divides 4D light filed into different views $(u,v)$ and corresponding images $(x,y)$. For a better understanding, we will show LFSP and graph structure in a multi-view way.

Super-pixel is a small patch composed of a series of pixels with similar appearance, brightness, texture in the 2D image, losing the structure information of scene. However, LFSP is composed of light rays emitted from a small region of a real object with similar characteristics. Light rays can be influenced by the structure of scene. That is to say, LFSP has a physical meaning. Supposing there is a light field recorded as $L(u,v,x,y)$, and R is a region of 3D space with similar features.
Mathematically, the LFSP $S_R (u,v,x,y)$ could be defined as,
\begin{equation}
\small
{S_R}\left( {u,v,x,y} \right) = \bigcup\limits_{i = 1}^{\left| R \right|} {L\left( {u_{P_i},v_{P_i},x_{P_i},y_{P_i}} \right)},
\end{equation}
where $|R|$ denotes the number of elements at region $R$ and $L({u_{P_i},v_{P_i},x_{P_i},y_{P_i}})$ is a bundle of rays which are emitted from the point ${P_i}$ in region $R$ \cite{zhu20174d}.

Obviously, LFSP is 4-dimensional and consists of angular dimensions $(u,v)$ and spatial dimensions $(x,y)$. In each view $(u^*,v^*)$, there is a patch ${S_R^{u^*,v^*}}$ corresponding to a 3D object. Because LFSP represents a region of a 3D object, the appearance of patches ${S_R^{u^*,v^*}}$ in different viewpoints are similar, which is called self-similarity of LFSP. It is remarkable that the redundancy of light field in angular dimension is guaranteed by LFSP. By fixing two dimensions $(u,x)$ to $(u^*,x^*)$ we can obtain EPI image of light field. LFSP is a combination of similar EPI lines. EPI lines with similar gradient and appearance make up LFSP.

\subsection{Graph structure based on LFSP}

Different from traditional 2D image, for each pixel $p=(u,v,x,y)$ in light field, there are two kinds of neighborhoods, the spatial and the angular, which can be formulated as,
\begin{equation}
\small
\begin{array}{l}
{N_{spa}}(p) = \left\{ \begin{array}{l}
(u,v,x \pm 1,y + 1)\\
(u,v,x \pm 1,y - 1)\\
(u,v,x \pm 1,y)\\
(u,v,x,y \pm 1)
\end{array} \right.\\
{N_{ang}}(p) = \left\{ \begin{array}{l}
(u \pm 1,v + 1,x \pm d(p),y + d(p))\\
(u \pm 1,v - 1,x \pm d(p),y - d(p))\\
(u \pm 1,v,x \pm d(p),y)\\
(u,v \pm 1,x,y \pm d(p)),
\end{array} \right.
\end{array}
\end{equation}
where ${N_{spa}}(p)$ and ${N_{ang}}(p)$ represent spatial and angular neighbors respectively, $d(p)$ represents the disparity. Since the shape of LFSP is irregular in most cases, the adjacent relationship of LFSP needs to be redefined. Let $M$ denotes the size of LFSP, ${S_i^{u,v}}$ and ${S_j^{u,v}}$ denote the 2D slices of LFSP $S_i$ and $S_j$ on view $(u,v)$ respectively. Then the spatial relationship of two LFSPs could be defined as,
\begin{equation}
\small
{N_{spa}}({S_i^{u,v}},{S_j^{u,v}}) = \left\{ \begin{array}{l}
1{\mbox{\quad if\quad}}\left\| {{P_{x,y}}(S_i^{u,v}) - {P_{x,y}}(S_j^{u,v})} \right\| \le \sqrt 2 M\\
0{\mbox{\quad otherwise}},
\end{array} \right.
\end{equation}
where $\left\| {\cdot} \right\|$ is the Euclidean distance between the center of $S_i^{u,v}$ and $S_j^{u,v}$. For angular relationship of LFSP, there are two kinds of angular neighbors, direct and indirect. Slices of one LFSP at different viewpoints are the direct angular neighbors of the LFSP. Due to the self-similarity of LFSP, the direct angular neighbors are similar in shape, color and position. Supposing $S_i$ and $S_j$ are spatial neighbors in viewpoint $(u,v)$, and $S_j^{u,v}$ and $S_j^{u^*,v^*}$ are direct angular neighbors, then indirect angle adjacent can be defined as,
\begin{equation}
\small
{N_{ang}}(S_i^{u,v},S_j^{{u^*},{v^*}}) = \left\{ \begin{array}{l}
1{\mbox{\quad }}{\exists (u,v)} \quad {N_{spa}}({S_i^{u,v}},{S_j^{u,v}}) = 1  \\
0{\mbox{\quad otherwise}},
\end{array} \right.
\end{equation}

In most cases, if LFSP $S_i$ and $S_j$ are spatial neighbors in viewpoint $(u,v)$, they are very likely to be spatial neighbors in viewpoint $(u^*,v^*)$. Considering the special cases that there are new spatial neighbors in viewpoint $(u^*,v^*)$, we introduce the indirect angle adjacent to assess this condition.

\begin{figure}[htb]
  \centering
  \includegraphics[width=\linewidth]{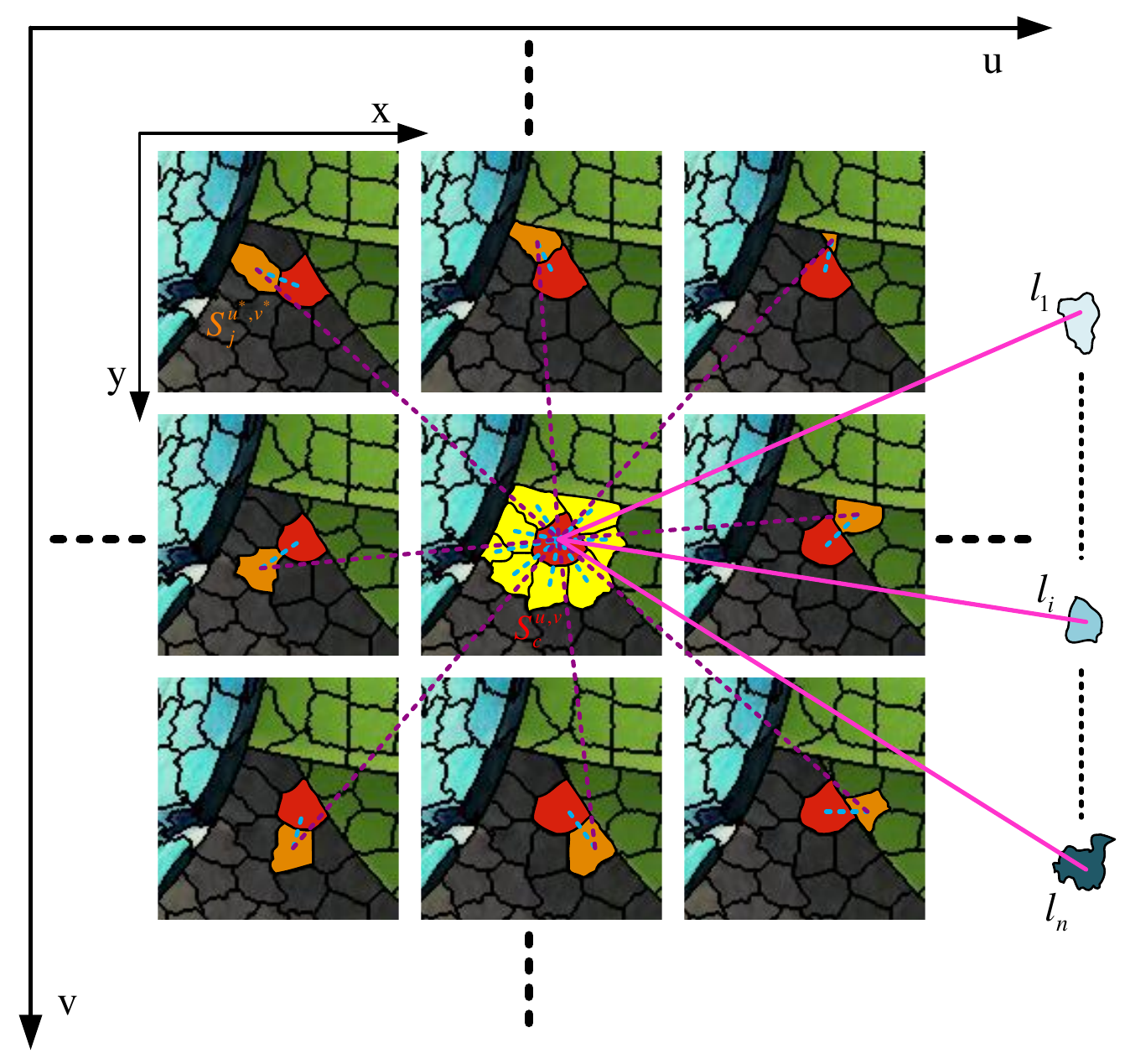}
  %
  %
  \caption{4D graph structure based on LFSP.}
  \label{fig:im4}
\end{figure}

After defining the neighboring relationship between LFSPs, we build the graph $G = ({\nu ,\varepsilon })$ of 4D light field based on LFSP. The vertex $\nu$ is composed by non-labeled LFSPs $N_S$ and labeled LFSPs $L_S$. $L_S$ is terminal of graph and is initialized by propagating user inputs to LFSP. The edge $\varepsilon$ is defined by spatial adjacent and angular adjacent,
\begin{equation}
\small
\left\{ \begin{aligned}
\nu  &= {N_S} \cup {L_S}\\
\varepsilon  &= {N_{spa}} \cup {N_{ang}},
\end{aligned} \right.
\end{equation}
The graph structure in a multi-view representation is shown in Figure \ref{fig:im4}, where each picture is a partial magnification of LFSP slices. In order to facilitate the representation, we choose $3 \times 3$ views instead of all views. In central view image, a LFSP $S_c^{u,v}$ is represented by a red patch and its spatial neighbours are marked in yellow. There are red patches in other views, which are the direct angular neighbours. In addition, there are orange LFSPs $S_j^{{u^*},{v^*}}$ next to the red LFSP, defined as indirect angular neighbours of $S_c^{u,v}$. The right side are the labeled LFSP $L_S$, in a gradually darken cyan color. The pink lines connect $S_c^{u,v}$ and $L_S$, representing the similarity between them. The blue dotted lines represent the influence of spatial neighbors. The purple dotted lines connect $S_c^{u,v}$ and $S_j^{{u^*},{v^*}}$, representing the influence of indirect angular neighbors.

\section{Energy Function}

After building the graph structure, the 4D light field segmentation is treated as an energy minimization problem. Supposing $L$ is a label vector that assigns label $l_{S_i}$ to LFSP $S_i$, the energy function is defined as follows,
\begin{equation}
\small
\begin{aligned}
E(L)\!=\!\!\!\!\!\!\sum\limits_{\tiny\begin{array}{c}{S_i} \in {N_S}\\{S_l} \in {L_S}\end{array}} {\left( {{E_C}({S_i},{S_l}) + {\lambda _p}{E_P}({S_i},{S_l}) + {\lambda _d}{E_D}({S_i},{S_l})} \right)}\\
	    +{\lambda_s} \sum \limits_{\{{S_i},{S_n}\} \in {N_{spa}}}E_{S}({S_i},{S_n})
	   + {\lambda_a} \sum \limits_{\{{S_i},{S_a}\} \in {N_{ang}}}E_{A}({S_i},{S_a}),
\end{aligned}
\end{equation}
where $E_C$, $E_P$ and $E_D$ measure the color, position and disparity distance between the non-labeled LFSP $S_i$ and the labeled LFSP $S_l$. $E_S$ and $E_A$ measure the connectivity between the adjacent LFSPs in spatial and angular space respectively. We give the detailed definition of energy function terms in the following section.



\subsection{Data term}

As an interactive segmentation, a user needs to draw different scribbles over the interested objects on the reference view of light field. In this paper, we choose central view to draw scribbles for ease of calculation, and other view can be chose by propagating disparity from central view to corresponding view. Then we propagate user inputs to LFSP, getting the labeled LFSP $L_S$. As seeds, $L_S$ is utilized to calculate the similarity with non-labeled LFSP $N_S$, using color, disparity and position cues.

Color, disparity and position represent different physical characteristics of objects. So it is important to integrate these cues for light field segmentation. There are some effective classifiers such as SVMs \cite{hearst1998support} or Random forests \cite{gislason2006random}. When the users are interested in a particular object or want to edit light field, they will frequently change the input scribbles. In that case, we calculate the similarity of color, disparity and position individually and use weights to balance influence of different cues. Experimental results show that our simple classifier can work effectively.

The color, disparity and position energy term are defined as,
\begin{equation}
\small
\begin{aligned}
{E_C}\left( {{S_i},{S_l}} \right)& = Normalize\left(\sum {\left\| {{C_{cha}}(S_i^{u,v}) - {C_{cha}}(S_l^{u,v})} \right\|} \right)\\
{E_P}\left( {{S_i},{S_l}} \right)& = Normalize(\left\| {{P_{x,y}}(S_i^{u,v}) - {P_{x,y}}(S_l^{u,v})} \right\|)\\
{E_D}\left( {{S_i},{S_l}} \right)& = Normalize(\left\| {D(S_i^{u,v}) - D(S_l^{u,v})} \right\|),
\end{aligned}
\end{equation}
where $C_{cha}(S_i^{u,v})$, ${P_{x,y}}(S_i^{u,v})$ and $D(S_i^{u,v})$ denote color, position and disparity information of LFSP $S_i$ respectively. $\sum $ combines influence of different channels. $Normalize$ is a normalization function changing the value to $[0,1]$. In our experiments, LFSPs are the basic unit of processing and need to be initialized. The color and disparity of a LFSP $S_i$ are the mean value of its components, and be defined as,
\begin{equation}
\small
\begin{aligned}
{C_{cha}}(S_i^{u,v}) &= \frac{1}{{\left| {S_i^{u,v}} \right|}}\sum\limits_{p \in S_i^{u,v}} {{C_{cha}}} (p)\\
D(S_i^{u,v}) &= \frac{1}{{\left| {S_i^{u,v}} \right|}}\sum\limits_{p \in S_i^{u,v}} D (p)\\
{P_{x,y}}(S_i^{u,v}) &= \frac{1}{{\left| {S_i^{u,v}} \right|}}\sum\limits_{p \in S_i^{u,v}} {{P_{x,y}}} (p) ,
\end{aligned}
\end{equation}
where ${C_{cha}}$ represents the channels of color space (here we use $CIELab$ color space). $p$ is the pixel belonging to $S_i$ in $(u,v)$ view.

\subsection{Smooth term}

When building graph structure, we have already mentioned the two adjacent relationship of light field. Spatial neighbors and angle neighbors act on smooth term, which has a great influence on the optimization result. Specifically, spatial neighbors encourage segmentation to be regular in the 2D slice $u,v$ of light field, while angle neighbors ensure consistency between the result in different view.

The definition of smooth term for spatial neighbors is as follows,
\begin{equation}
\small
E_{S}({S_i},{S_n})= \delta \left( {{l_{{S_i}}},{l_{{S_n}}}} \right) {B_{{S_i},{S_n}}},
\end{equation}
where $\delta$ is a penalty factor of two LFSPs.
\begin{equation}
\small
\delta ({l_{{S_i}}},{l_{{S_n}}}) = \left\{ \begin{array}{l}
0, \qquad if \quad {l_{{S_i}}} = {l_{{S_n}}}\\
1, \qquad otherwise ,
\end{array} \right.
\end{equation}
When the labels are not the same, a penalty is added. 

${B_{{S_i},{S_n}}}$ is the LFSP similarity between $S_i$ and $S_n$, which is evaluated by,
\begin{equation}
\small
\begin{aligned}
{B_{{S_i},{S_n}}} &= exp\left( { - \frac{{\sum {\left\| \triangle C_{cha} \right\|} }}{{\sigma _C^2}}
-\alpha \frac{{\left\| \triangle D \right\|}}{{\sigma _D^2}}} \right)\\
\triangle C_{cha}&=C_{cha}(S_i) - C_{cha}(S_n),\triangle D=D({S_i}) - D({S_n}),
\end{aligned}
\end{equation}
where $\sigma _C^2$ and $\sigma _D^2$ are variance of color and disparity, $\alpha$ is used to balance the influence of two cues.Position cue is not used because adjacent relationship is determined by position information. The smooth term from all angle neighbors is defined as,
\begin{equation}
\small
E_{A}({S_i},{S_a}) = \delta \left( {{l_{{S_i}}},{l_{{S_a}}}} \right) A({S_a}) {B_{{S_i},{S_a}}}  ,
\end{equation}
where $A({S_a})$ is used to determine whether ${S_a}$ is new angle neighbor of ${S_i}$. If ${S_a}$ is not a new angle neighbor, it will have a repetitive effect on the smoothing processing, which is not allowed. Concretely,
\begin{equation}
\small
A({S_a}) = \left\{ \begin{array}{l}
1 \qquad {\mbox{if }}{{S_a}}{\mbox{ is a new angle neighbor}}\\
0 \qquad {\mbox{otherwise}} ,
\end{array} \right.
\end{equation}


The $\alpha$-expansion algorithm \cite{Boykov2002Fast} is used to solve Eqn.(6). The complete process of our algorithm is shown in Algo.1. 4D light field and user scribbles are the input of our algorithm. First, disparity map in central view is calculated by \cite{zhu2017occlusion}. Then LFSP is obtained by algorithm \cite{zhu20174d}. And then LFSP information is initialized, getting number, color disparity of each LFSP. At the same time, user inputs are propagated to LFSP and some LFSP are labelled as seeds in the 4D graph. Next, the penalization of giving non-labeled $S_i$ a label $l$ is calculated by $E_{data}$, meanwhile the influence of neighbors $S_i$ and $S_n$, $S_a$ is calculated by $E_{smooth}$. We get total energy $E(L)$ and use a graph cut algorithm to minimize the energy function. Finally, we obtain the optimized segmentation result.

\begin{algorithm}[htbp]\label{algo:full_algo_description}
\small
\SetAlFnt{\small\sf}
\SetAlCapFnt{\small}\SetAlCapNameFnt{\small}
\caption{Light field segmentation algorithm based on LFSP graph representation.}
\small
\KwIn{The 4D light field \textit{LF} and input label \textit{Inlabel}}
\KwOut{The 4D LF segmentation result \(seg\).}

\(D = DepthEstimation(LF)\)

\(LFSP = GetLFSP(LF)\)

\For{all LFSP}{	
	${{C_{cha}}(S_i^{u,v}) = \frac{1}{{\left| {S_i^{u,v}} \right|}}\sum\limits_{p \in S_i^{u,v}} {{C_{cha}}} (p)}$
	
	${D(S_i^{u,v}) = \frac{1}{{\left| {S_i^{u,v}} \right|}}\sum\limits_{p \in S_i^{u,v}} D (p)}$
	
	${{P_{x,y}}(S_i^{u,v}) = \frac{1}{{\left| {S_i^{u,v}} \right|}}\sum\limits_{p \in S_i^{u,v}} {{P_{x,y}}} (p)}$

}

\For {non-label LFSP  \({S_i}\) in central view \(({u_0},{v_0})\)}{
	\({E_{data}({S_i},{S_l})} = { {{E_C}({S_i},{S_l}) + {\lambda _p}{E_P}({S_i},{S_l}) + {\lambda _d}{E_D}({S_i},{S_l})} } \)
	
	\For{ neighbour \({{S_i},{S_n},{S_a}}\)}{

\(E_{smooth}({S_i}) = {\lambda _s}E_{S}({S_i},{S_n})
	   + {\lambda _a}E_{A}({S_i},{S_a})\)
	}
}

\(E(L) = \sum {{E_{data}}({S_i},{S_l})} + \sum {{E_{smooth}}({{S_i}})}\)

\(seg = {{\mbox {argmin}} _L}E(L)\)

\end{algorithm}

\section{Experiments}

In this section, we compare our segmentation algorithm with three state-of-art light field segmentation algorithms, including GCMLA (globally consistent multi-label assignment) \cite{Wanner2013Globally}, SACS (spatial and angular consistence segmentation) \cite{mihara20164D} and RBGSS (ray-based graph structure segmentation) \cite{hog2016light}. Synthetic data and real data are used to demonstrate the performance of our algorithm. For synthetic data, we use the HCI dataset proposed in \cite{wanner2013datasets}, which contains 4 light fields with known depth, ground truth labeling and user input scribbles. Because there are few light field datasets with ground truth segmentation results, except the HCI benchmark. For real data, two popular light field camera Lytro and Illum are used to capture light field of scenes, and light field are decoded by LFToolbox \cite{Dansereau2013Decoding} and Lytro Power Tools respectively. We make comparative experiments on segmentation accuracy and running time. In addition, we also verify the validity of smoothing term. Finally, we analyze   the limiting cases of our algorithm. The code of GCMLA is implemented in $cocolib$ coming from author's website, and the result of SACS and RBGSS are obtained from their paper respectively.

\begin{figure*}[tbp]
  \centering
  \mbox{} \hfill
  \includegraphics[width=.8\linewidth]{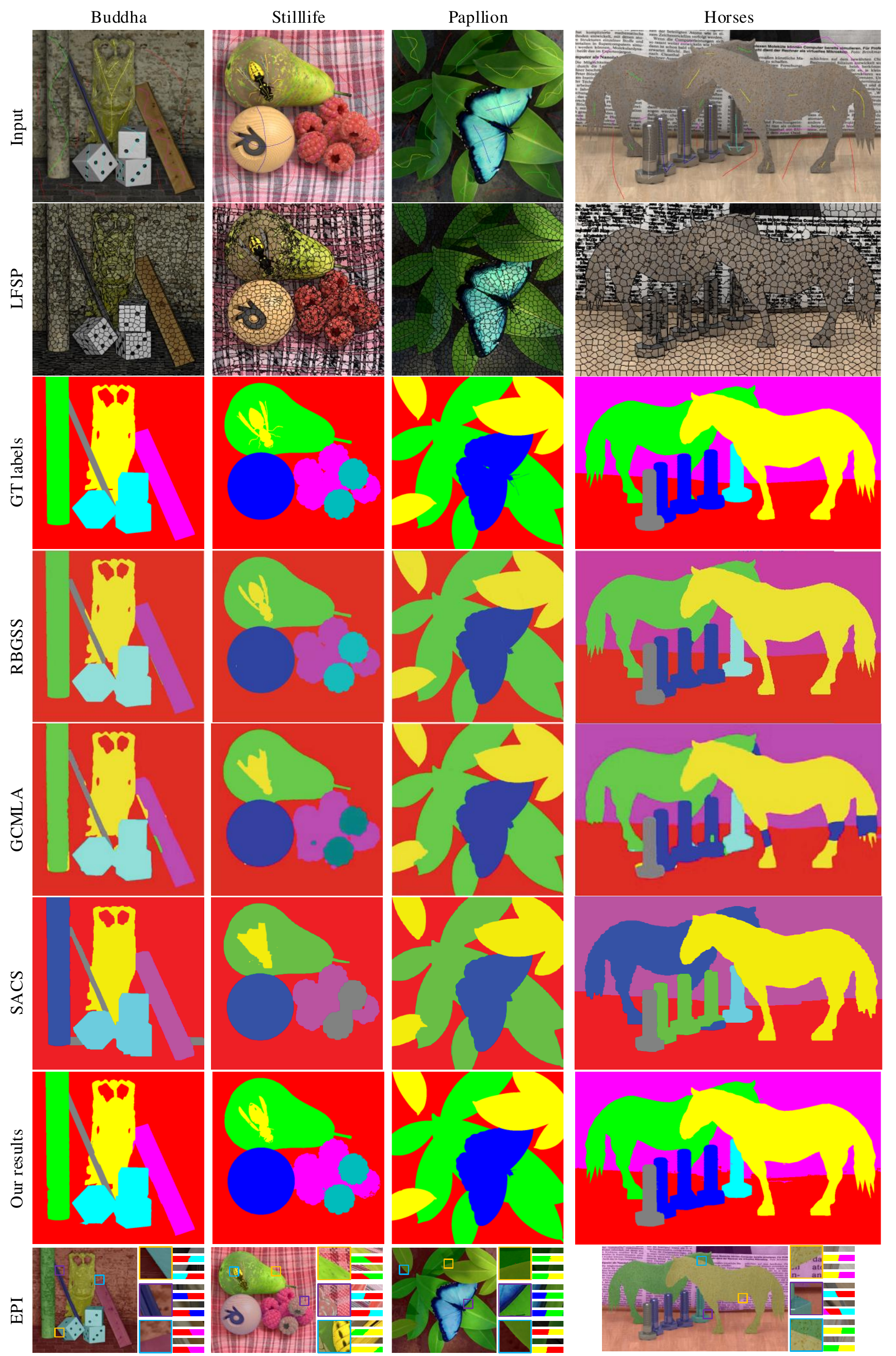}
  \hfill \mbox{}
  \caption{\label{fig:im7}%
           Light field segmentation results on synthetic datasets\cite{wanner2013datasets}.
           }
\end{figure*}
Parameter setting is important to the algorithm, and then we introduce the important parameter settings and their significance. In data term, $\lambda _p$ and $\lambda _d$ are used to balance the influence of color, position and disparity. Because $E_C$, $E_P$ and $E_D$ are normalized, it is reasonable to assign similar value to $\lambda _p$ and $\lambda _d$. Since the real data and its disparity are relatively noisy, $\lambda _d$ is set with a smaller value to provide some robustness. A larger $\lambda _p$ leads to a more regular result. $\lambda _s$ and $\lambda _a$ control the smoothing term to avoid over-smoothing. It is not recommended to assign a large value to $\lambda _s$ and $\lambda _a$ because initial results are good enough. During the experiment, $\lambda _s=10$ and $\lambda_a=2$, because spatial neighbor plays a dominate role and indirect angular neighbor are relatively rare.

\subsection{Synthetic data}

Figure \ref{fig:im7} shows the segmentation results of HCI dataset obtained by state-of-art light field segmentation algorithm. The first row displays the slices in central view respectively, and there are user's scribbles on the central view image. The second row displays LFSP slices in the central view. The third row is the segmentation ground truth in the central view provided by HCI dataset. The next few rows are the segmentation results of four algorithm (RBGSS, GCMLA, SACS and our algorithm) respectively. The last column is the EPI pictures of our segmentation results. The EPI picture shows that our algorithm could segment 4D light field and the segmentation result in different view is coherent. While several algorithms have similar performance, the results of our method are more accurate in the boundary and detailed areas. The quantitative analysis is provided in Table. 1 where the accuracy of each scene and the average accuracy are calculated. The ground truth and estimated depth are used to compare different methods. In order to enhance the contrast, we stress the best results with Bold Fonts. In most cases, our method outperforms all the other evaluated techniques. It is worth noting that our method achieves the highest average accuracy rate.

\begin{table*}
\begin{center}
\small
\begin{tabular}{c|cccc|ccc}
\hline
Algorithm & GCMLA & RBGSS         & Ours          & w/o smooth & GCMLA & SACS & Ours          \\ \hline
Depth     & GT    & GT            & GT            & GT         & EST   & EST  & EST           \\ \hline
Papillon  & 99.4  & \textbf{99.5} & \textbf{99.5} & 99.4       & 98.9  & 98.3 & \textbf{99.4} \\
Buddha    & 99.1  & 99.1          & \textbf{99.2} & 96.3       & 98.8  & 96.4 & \textbf{99.0} \\
Stilllife & \textbf{99.2}  & \textbf{99.2} & 99.1          & 98.2       & \textbf{98.9}  & 97.7 & \textbf{98.9} \\
Horses    & 99.1  & 99.1          & \textbf{99.3} & 99.1       & 98.3  & 95.9 & \textbf{98.9} \\ \hline
Average   & 99.2  & 99.2          & \textbf{99.3} & 98.3       & 98.7  & 97.1 & \textbf{99.0} \\ \hline
\end{tabular}
\end{center}
\caption{Light field segmentation accuracy comparisons using the percentage of correctly segmented pixels.}
\end{table*}

In addition to high accuracy, the significant advantage of our approach is computational efficiency. GCMLA only gets the segmentation result in the central view and the running time is long. SACS uses $5 \times 5$ views of $9 \times 9$ views to reduce the data size, but the algorithm complexity is high. By using ray-bundle and free rays, RBGSS reduces running time to some extent. They perform the optimisation in 4 to 6s on an Intel Xeon E5640. Our method greatly reduces data size and achieve a great promotion in computational efficiency. The preprocessing (depth estimation and light field super-pixel segmentation) of our method takes around 70s. While the average segmentation time is about 1.4s which is close to the requirements of real-time processing. Furthermore, the preprocessing and segmentation time could be achieved by GPU acceleration. Noting that, our algorithm is evaluated on a desktop computer with a 3.6 GHz i7 CPU.

Supposing there is a light field with $9\!\times\!9$ views and the size of LFSP is $20 \times 20$, theoretically our method can simplify the data size by $81\!\times\!400$ times. Specifically, taking Buddha data as an example, \cite{hog2016light} can reduce data size from $4.77\!\times\!10^7$ to $8.19\!\times\!10^5$, while our method can reduce data size from $4.77\!\times\!10^7$ to $1.46\!\times\!10^3$. \cite{Wanner2013Globally,mihara20164D}  can not reduce data size in their algorithms so that they are time-consuming.

\begin{figure}[htb]
  \centering
  \includegraphics[width=\linewidth]{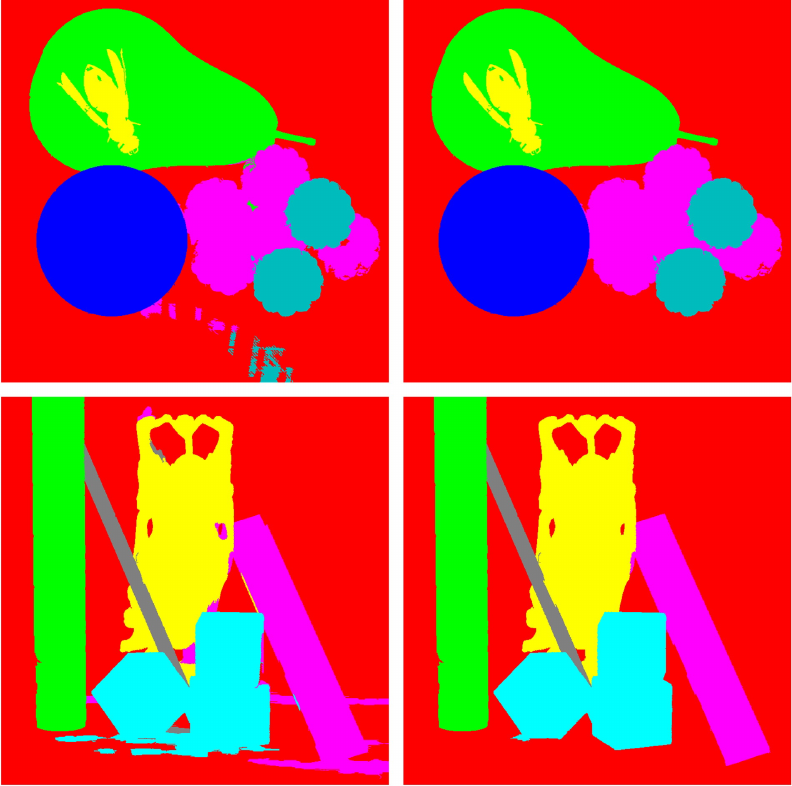}
  %
  %
  \mbox{(a)        \hspace{38mm}               (b)}\hfil
  \caption{\label{fig:im5}
           Effectiveness of smooth term. (a) Initial segmentation results. (b) Optimized segmentation results.}
\end{figure}

Furthermore, we evaluate the effectiveness of two smooth constraints, spatial adjacent and angle adjacent. Portion of the results are shown in Figure \ref{fig:im5}. The right picture is optimized result with smooth constraints and the left one is initial segmentation result without smooth constraints. The initial segmentation result is noisy and suffers from a high error rate. The optimized result is more accurate for both central view image and 4D light field, proving the validity of two smooth constraints.

\subsection{Real data}

We use multiple sets of real light field data to verify the effectiveness of our algorithm. For each set of real data, we show its LFSP in central view image, user input, disparity map, segmentation result in central view and EPI result respectively. The disparity map for Illum can be obtained from the camera. For Lytro the disparity map should be estimated by other's work (Here we use the algorithm proposed by \cite{zhu2017occlusion} ). The EPI results show accuracy and consistency of segmentation result across all views.

\begin{figure*}[tbp]
  \centering
  \mbox{} \hfill
  \includegraphics[width=1\linewidth]{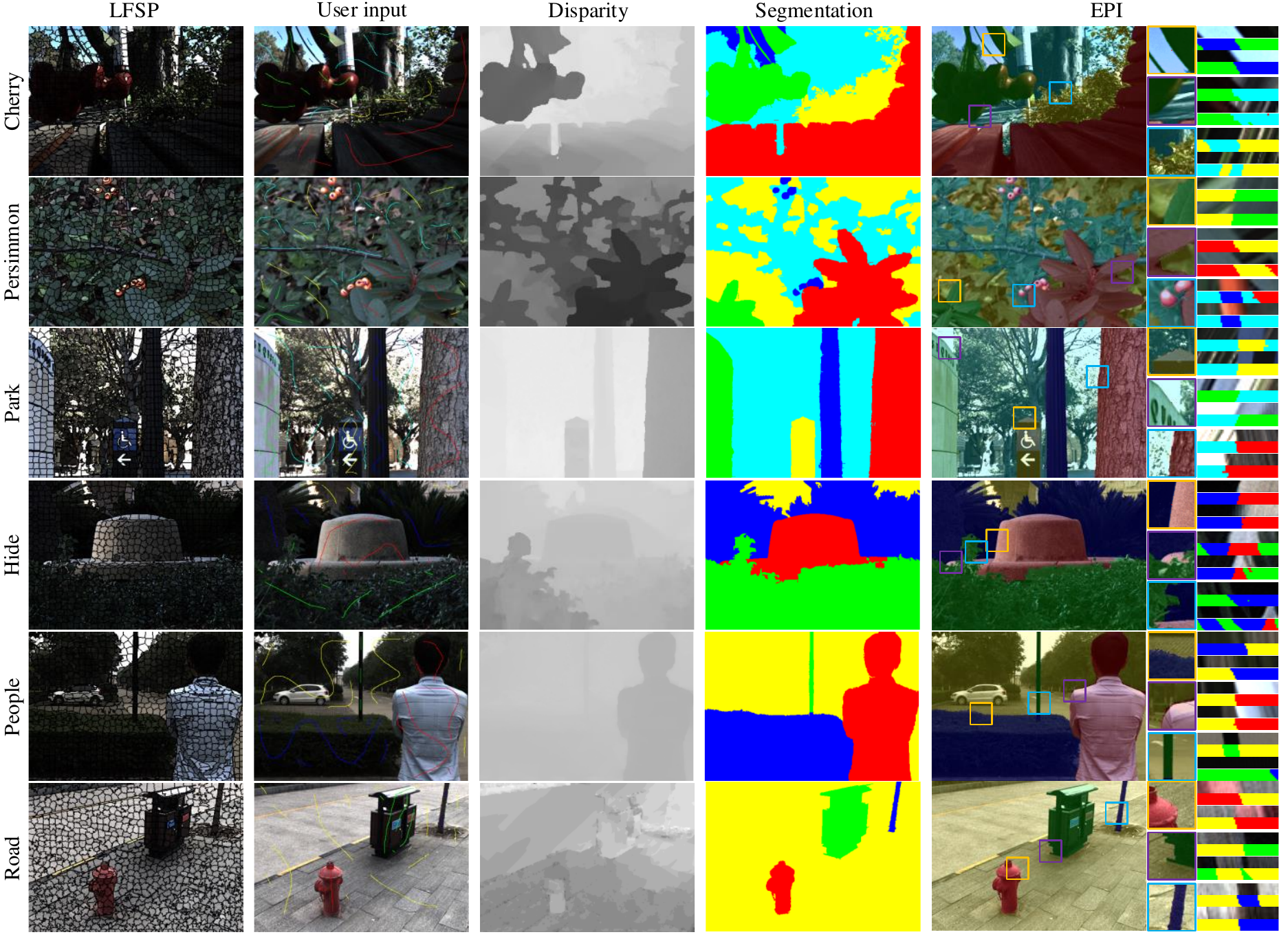}
  \hfill \mbox{}
  \caption{\label{fig:im8}%
           Segmentation results for real data captured by Illum. The first four sets of data (Cherry, Persimmon, Park and Hide) are provided by \cite{rajlight}. The last two sets of data (People, Road) are captured by our Illum camera. EPI images consist of overlap image, local amplification images, horizontal EPI of raw image, horizontal EPI of segmentation results, vertical EPI of raw image and vertical EPI of segmentation results respectively.}
\end{figure*}

\begin{figure*}[tbp]
  \centering
  \mbox{} \hfill
  \includegraphics[width=1\linewidth]{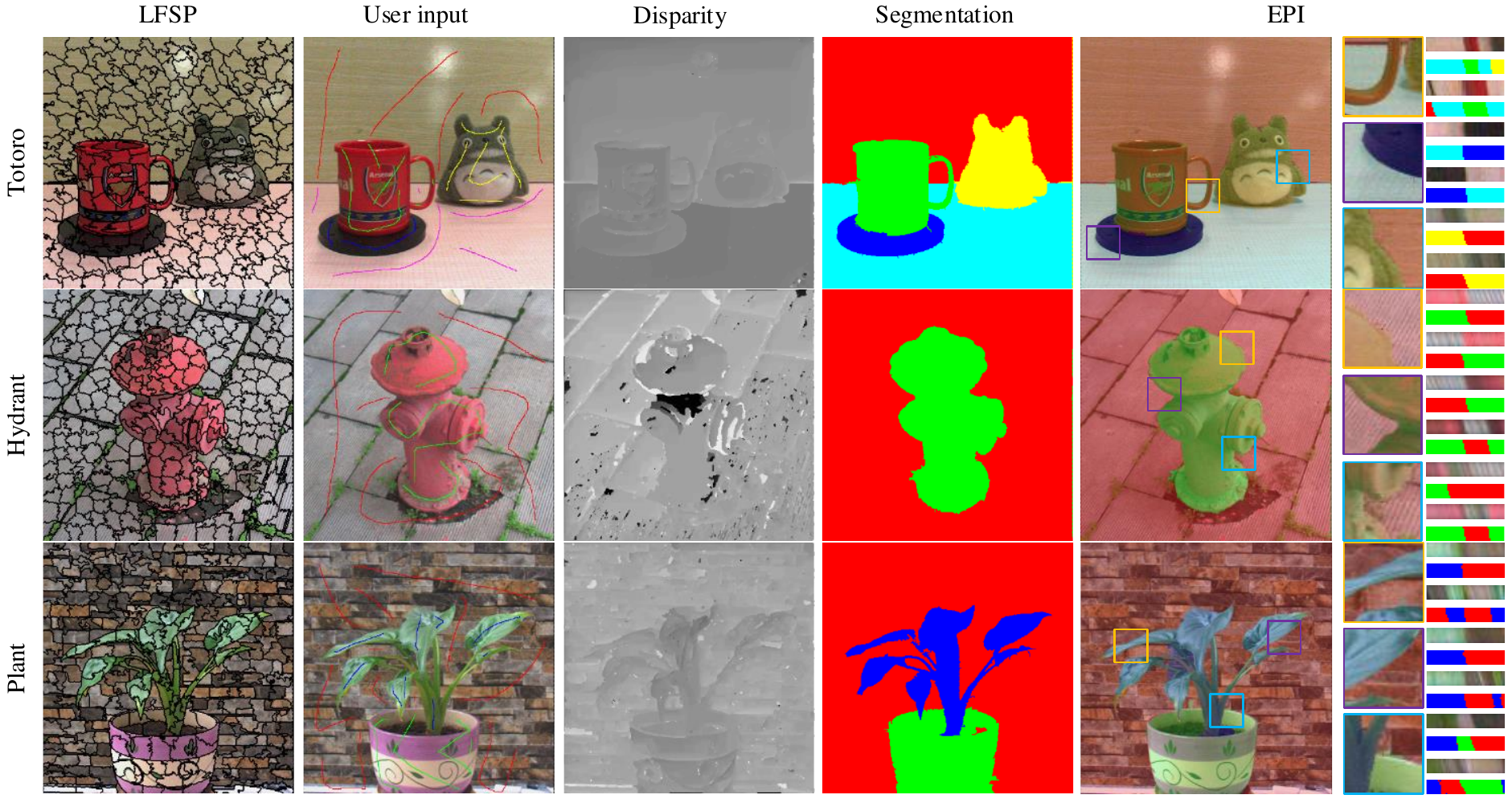}
  \hfill \mbox{}
  \caption{\label{fig:im9}%
           Segmentation results for real data captured by Lytro. EPI images consist of overlap image, local amplification images, horizontal EPI of raw image, horizontal EPI of segmentation results, vertical EPI of raw image and vertical EPI of segmentation results respectively.}
\end{figure*}

\begin{figure*}[tbp]
  \centering
  \mbox{} \hfill
  \includegraphics[width=1\linewidth]{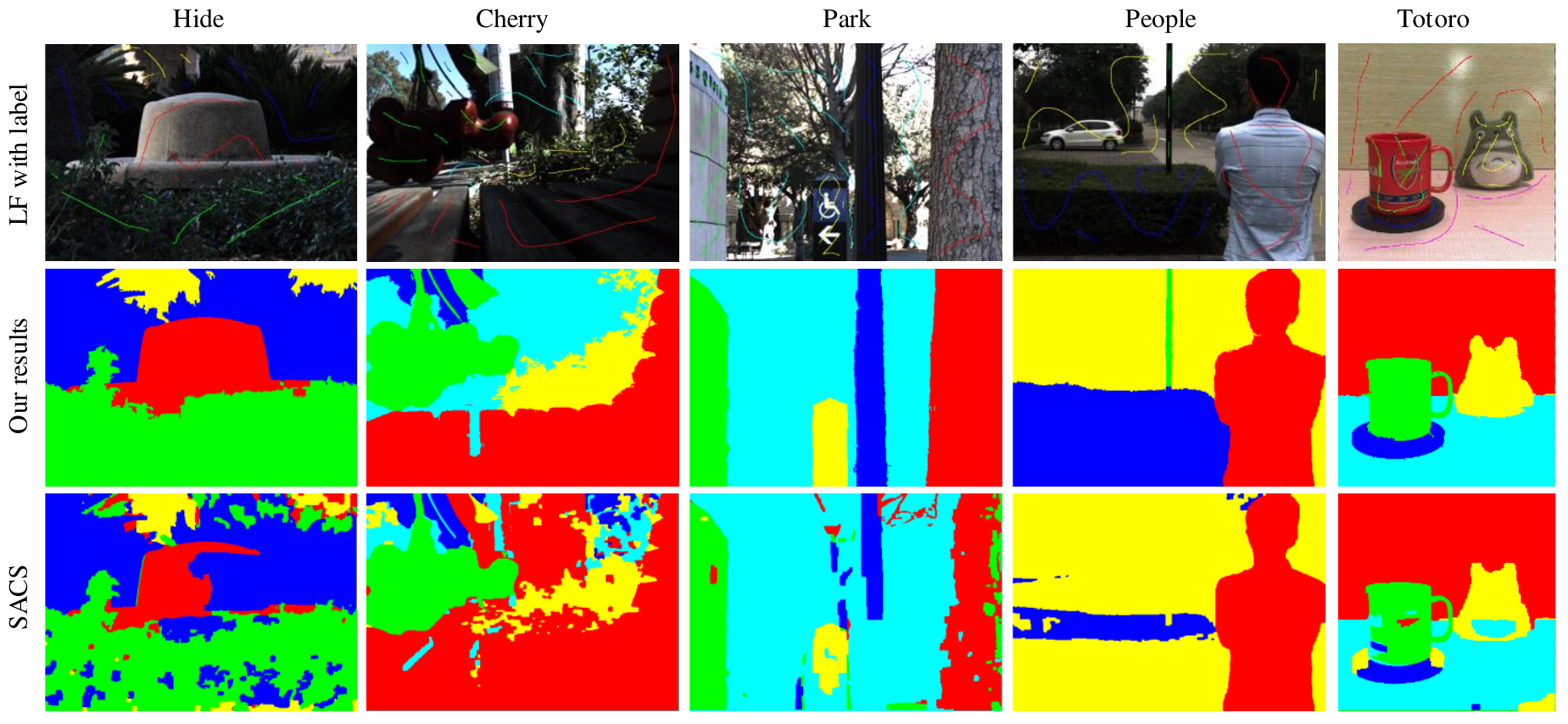}
  \hfill \mbox{}
  \caption{\label{fig:im10}%
           Comparison of real data segmentation results of different algorithms. The first row shows central view images with scribbles. The second row is segmentation results of our method and the last row is segmentation results of \cite{mihara20164D}.}
\end{figure*}

Figure \ref{fig:im8} shows the segmentation results of real data captured by Illum. Light field data from `Cherry' to `Hide' are provided by \cite{rajlight}. `People' and `Road' are captured by our Illum camera. Figure \ref{fig:im9} shows segmentation results of real data captured by Lytro. Due to the low quality of light field data, the disparity map is poor and noisy, especially for Lytro data. So the segmentation algorithm ought to be robust against noise and errors. RBGSS is sensitive to depth map quality because ray-bundles and free rays are obtained by depth information. However, our method is robust to depth map quality, as shown in the segmentation and EPI results. Figure \ref{fig:im10} shows the comparison of real data segmentation of different algorithms. When segmenting real data of light field, we reduce the value of disparity weight to decrease the influence of depth map errors. So we can see that the segmentation results are fine.

In addition, the proposed method is a complete 4D light field segmentation. As shown in Figure \ref{fig:im1} and Figure \ref{fig:im8}, the EPI lines of raw data and segmentation result are consistent, which demonstrates the validity of the proposed algorithm in different views. The coherent segmentation across all views is available, which is important for light field editing. Moreover, the segmentation result can be used to change the color of specific region, remove occluding object from a scene and so on.

\subsection{Limitations}

\begin{figure}[htb]
  \centering
  \includegraphics[width=1\linewidth]{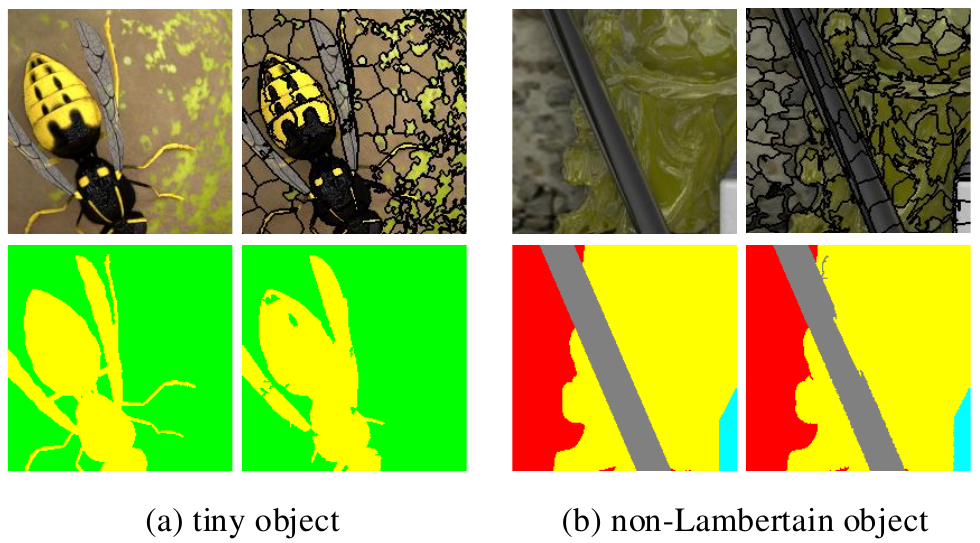}
  %
  %
  \caption{\label{fig:im6}
  Imperfect cases of our segmentation algorithm caused by (a) tiny objects and (b) non-Lambertian regions. Each part consists the enlargement of raw image, LFSP, ground truth labels and segmentation results respectively. }
\end{figure}

There are some limitations of our algorithm. The segmentation results are affected by LFSP quality to some extent. For example, when some tiny objects are similar to background, or a non-Lambertian object reflects the ray of adjacent objects, in such cases, our method is likely to have a poor performance. Figure \ref{fig:im6} shows two limited situations. In figure (a), some tiny objects (feet and wings of bees) have similar or the same texture characteristics as the background. It is difficult to distinguish them from the background. In figure (b), a smooth glass column reflects the color of Buddha, making segmentation boundary confused. These two limitations are tough problems of image segmentation which are not solved well so far.

\section{Conclusions}

In this paper, we propose to utilize LFSP to interactively segment light field. The characteristics of LFSP in spatial and angular domains are helpful to improve segmentation accuracy and efficiency. We propose a novel 4D graph structure based on LFSP and define the spatial and the angular neighbors accordingly. Then a data term and smooth one of energy function are defined according to graph structure. After that, a graph cut algorithm is used to optimize segmentation result. Experiments on synthetic data show that the proposed method not only has high accuracy, but also has high computational efficiency compared to state-of-the-art algorithms. Moreover, we apply our method to real light fields, which shows the effectiveness and robustness of our algorithm. In the future, we will adapt the proposed method for light field video segmentation.

\section*{Acknowledgement}

We thank H. Mihara and T. Karin for their helps on real scene light field segmentation.

{\small
\bibliographystyle{ieee}
\bibliography{egbib}
}

\end{document}